\documentclass[journal]{IEEEtran}
\ifCLASSINFOpdf
\else
\fi
\usepackage{supertabular}
\usepackage{lscape}
\usepackage{amssymb}
\usepackage{graphicx,subfigure,amsmath}
\usepackage[usestackEOL]{stackengine}
\usepackage{multirow}
\usepackage{enumitem}
\usepackage{titlesec}
\usepackage{tabularx}
\usepackage{graphicx}
\usepackage{adjustbox}
\usepackage{rotating}
\usepackage{algorithm}
\usepackage{makecell}
\usepackage[table,xcdraw]{xcolor}
\usepackage{algpseudocode}
\usepackage{url}

\usepackage[numbers]{natbib}

\begin{document}
%
\title{CMED: A Child Micro-Expression Dataset}
%
%
%

\author{Nikin~Matharaarachchi,
        Muhammad~Fermi Pasha,
        Sonya~Coleman,
        and~Kah Peng~Wong 
}

%
%

\markboth{Pre-print submitted to IEEE Transactions on Affective Computing }%
{Shell \MakeLowercase{\textit{et al.}}: Bare Demo of IEEEtran.cls for IEEE Journals}
%



\maketitle

\begin{abstract}
Micro-expressions are short bursts of emotion that are difficult to hide. Their detection in children is an important cue to assist psychotherapists in conducting better therapy. However, existing research on the detection of micro-expressions has focused on adults, whose expressions differ in their characteristics from those of children. The lack of research is a direct consequence of the lack of a child-based micro-expressions dataset as it is much more challenging to capture children's facial expressions due to the lack of predictability and controllability. This study compiles a dataset of spontaneous child micro-expression videos, the first of its kind, to the best of the authors knowledge. The dataset is captured in the wild using video conferencing software. This dataset enables us to then explore key features and differences between adult and child micro-expressions. This study also establishes a baseline for the automated spotting and recognition of micro-expressions in children using three approaches comprising of hand-created and learning-based approaches.
\end{abstract}

\begin{IEEEkeywords}
micro-expressions, emotion, facial expression, dataset, baseline.
\end{IEEEkeywords}

%
\IEEEpeerreviewmaketitle

\section{Introduction}
%
%
%
%
\IEEEPARstart{M}{icro-expressions} (ME) are brief, spontaneous expressions that occur when someone suppresses an emotion. They are very difficult to control or hide. Micro-expressions were first discovered by Haggard and Isaacs\cite{Haggard1966}. They were greatly expanded by psychologist Paul Ekman\cite{Ekman1969}, who identified six basic emotions that could be expressed through micro-expressions: anger, disgust, fear, happiness, sadness, and surprise. 

Studies have shown that, from childhood, we learn that specific emotional responses conveyed through facial expressions are appropriate in a given situation. Thus, humans have been trained to hide their genuine emotions in situations due to social expectations or to deceive others deliberately \cite{Garner1996}. Owing to the short-lasting and involuntary nature of micro-expressions, even when people try to hide their genuine emotions by forcibly altering their macro-expressions, such as through the use of smiles, micro-expressions still present \cite{Porter2012b}. The ability to hide true emotions by forcing fake facial expressions has been studied widely. It has shown that the ability to judge the authenticity of emotions is vital in many fields such as education \cite{Chiu2013}, forensic investigations \cite{Davis2006}, health \cite{Lautenbacher}\cite{Marsh2012}, immigration, customs and security checks \cite{Sanchez-Monedero2019}\cite{Blandon-Gitlin2014}, court rulings \cite{Porter2009}, business \cite{Grandey2005}, politics \cite{Rossini2011}, robotics and human-computer interactions \cite{Bruce2002}.

Child emotions differ fundamentally from adults due to psychological and physiological differences as shown in previous psychological studies \cite{khosla2024understanding}. Mandy et al. \cite{visser2014contextual} have studied differences between adults and children in expressing surprise and find that children have more pronounced brow and mouth movements due to developmental differences and the lack of emotion regulation. In addition children are prone to rapid emotional changes. Furthermore, collecting samples from children comes with additional challenges due to the lack of predictability and controllability.

In the context of micro-expression detection in children, it benefits psychotherapists in performing technology-enhanced therapy. This can be especially beneficial for children who struggle to verbalize their emotions, such as those with autism spectrum disorder (ASD), anxiety, or trauma. In addition, automated micro-expression detection systems integrated into teletherapy platforms could allow therapists to monitor emotions even in remote settings. This is particularly valuable for children in rural areas or during situations where in-person sessions may not be feasible. However, to the best of our knowledge, there has been no attempt to create a dataset for child micro-expressions, and all samples in the primary micro-expression datasets are from adult subjects; thus, work on child micro-expression recognition is non-existent yet high beneficial. 

While micro-expressions in children may not have been widely studied, the number of studies on micro-expression recognition in adults has been increasing rapidly. Most of the early research has used hand-designed models for feature extraction \cite{ Fasel2003}. They can be classified into geometric-based features such as Main Directional Mean Optical Flow (MDMO)\cite{ Liu2016} and appearance-based features such as Local Binary Pattern- Three Orthogonal Planes (LBP-TOP)\cite{Li2018}, and 3D Histogram of Oriented Gradients (3D-HOG) \cite{Dalal2005}. However, with the advances in deep learning, recent studies focus on using various learning-based approaches to recognize micro-expressions.  The primary approach initially was to use a CNN and provide inputs such as a single frame \cite{ Patel2016} such as the apex frame \cite{ Li2018b}  or a sequence of interpolated frames. Studies have also been performed using pre-processed features such as optical flow \cite{ Liong}, optical strain  \cite{ Liong2019} and synthetic images \cite{ let2020} as inputs. More recently, graph-based approaches have been gaining popularity using its variants such as DGCNNs\cite{nikin} and T-GCNs\cite{mergcn}.

\section{Existing Micro-Expression Datasets}

Micro-expression recognition is the classification of a given set of frames (or single frame) consisting of a micro-expression from one of the emotion classes. While no micro-expression datasets exists for children, the primary datasets used for adult micro-expression recognition tasks are CASME \cite{ Yan2013a}, CASME II \cite{ Yan2014a}, CAS(ME)$^2$ \cite{ Qu2018}, SMIC \cite{ Li2013} and SAMM \cite{ Davison2018}. The number of samples present and the number of classes in each of the datasets can be found in Table \ref{datasets}.  In summary:

\begin{table}[!h]
\caption{Details of the existing micro-expression datasets.}
\label{datasets}
\begin{center}
{\scriptsize
\begin{tabular}{|l|l|l|l|l|}
\hline
\textbf{Dataset}  & \textbf{Subjects} & \textbf{Samples}   & \textbf{Classes} & \textbf{FPS} \\ \hline
\textbf{CASME}    & 35                & 195                & 8                & 60           \\ \hline
\textbf{CASME II} & 35                & 247                & 5                & 200          \\ \hline
\textbf{CAS(ME)$^2$ } & 22                & 53 (and 250 macro) & 4                & 30           \\ \hline
\textbf{SMIC}     & 16                & 164                & 3                & 100          \\ \hline
\textbf{SAMM}     & 32                & 159                & 7                & 200          \\ \hline
\textbf{CAS(ME)$^3$ } & 100                &  1109 (and 3190 macro)& 7                & 30           \\ \hline
\textbf{4DME} & 41                & 267 & 5                & 60           \\ \hline

\end{tabular}}
\end{center}
\end{table}

\begin{itemize}
    \item The Chinese Academy of Sciences Micro-expression (CASME) database 
 contains 195 micro-expressions filmed under 60 fps. The study included 195 samples from 35 subjects. The samples are divided into amusement, sadness, disgust, surprise, contempt, fear, repression, and tense.
 \item The Chinese Academy of Sciences Micro-expression II (CASMEII) database used higher temporal (200fps) and spatial resolution (about 280x340 pixels on facial area). It consisted of 247 micro-expressions from happiness, disgust, surprise, repression, and other.
 \item The Chinese Academy of Sciences Macro-Expressions and Micro-Expressions (CAS(ME)²) contains both macro-expressions and micro-expressions. It contains 53 expression samples that were recorded at 30 fps, and the resolution was set to 640 $\times$ 480 pixels. The samples were classed as Positive, Negative, Surprise and Other.
 \item The $\text{CAS(ME)}^{3}$ dataset consists of 1,109 samples that were classed as happiness, disgust, surprise, fear, anger, sadness, and other. The samples were collected in 30fps and a resolution of 1280 $\times$ 720 pixels
\item The Spontaneous Micro-Facial Movement Dataset (SAMM) consists of 159 micro-expressions, captured at a resolution of 2040 × 1088 pixels and 200 frames per second. A key feature of SAMM is its tailored emotional stimuli for each participant, inducing natural responses in a controlled lab setting. The samples were classed as anger, disgust, fear, happiness, sadness, surprise, and contempt.
\item The Spontaneous Micro-expression Database (SMIC) were collected from 16 participants at 100 fps and were classified into positive, negative and surprise. The dataset also contains a variant with 71 samples from 8 participants which were collected in 25 fps.
\item The 4D Micro-Expression Dataset (4DME) is a multimodal micro-expression database consisting of 267 micro-expressions and 123 macro-expressions from 41 participants and were classified as positive, negative, surprise, repression and others.
\end{itemize}

Some of the key limitations in the existing datasets in addition to the lack of child ME data are listed:

\begin{itemize}
    \item Limited dataset size: Most datasets have relatively small numbers of micro-expressions due to the difficulty of eliciting and capturing spontaneous facial movements.
    \item Challenges in inducing genuine micro-expressions: The controlled lab settings may not fully replicate real-world scenarios, potentially influencing the naturalness of the expressions. In addition, since the participants are made to elicit emotions based on videos shown, and are expecting it, the emotions expressed may be influenced.
    \item Devices used: While some of the datasets use standard hardware, some utilize specialized hardware such as high speed cameras which may limit the real-world applications of such data, as such hardware may not be readily available to be deployed in most situations.
\end{itemize}

These limitations along with the lack of child micro-expression data show the importance of creating a dataset dedicated to child micro-expression samples, and one that can aid the development of automatic micro-expression recognition systems that could be deployed in a wide variety of real-world scenarios. In addition to not requiring specialized hardware, by using a non-laboratory setting (in the wild), our dataset would be able to obtain samples with a greater degree of variety in environment such as lighting conditions and face angles. 

\section{Dataset Creation Methodology}
\begin{figure*}[h!]
\centering

    \includegraphics[width=1.9\columnwidth]{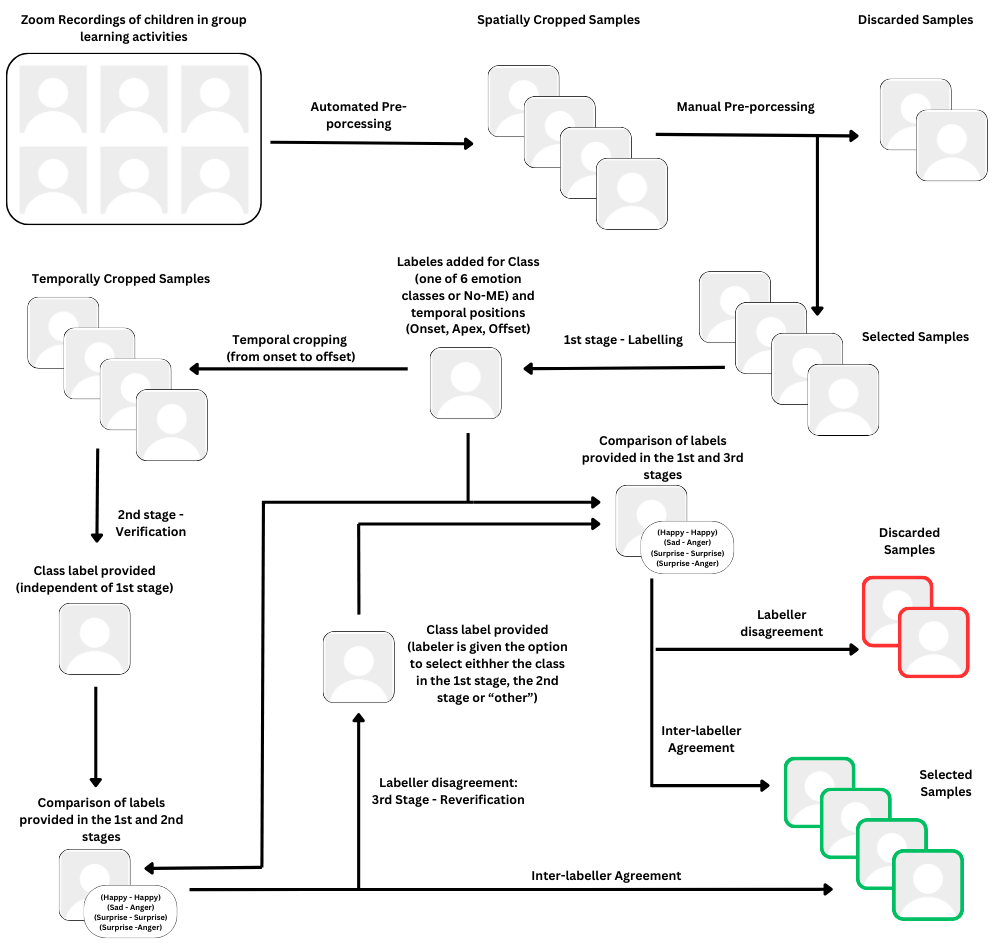}
    \caption{Dataset creation process }
\label{process}
\end{figure*}

In this section, we outline the method used to create the Child Micro-Expression Dataset (CMED). The overall process, as illustrated in Figure \ref{process}, starts by collecting data then processing it using both automated and manual approaches, then labelled using a three stage process, and finally the dataset is compiled. This section outlines each of the steps in the process. 

\subsection{Data Collection}
\begin{figure}[h]
\centering

    \includegraphics[width=.49\textwidth]{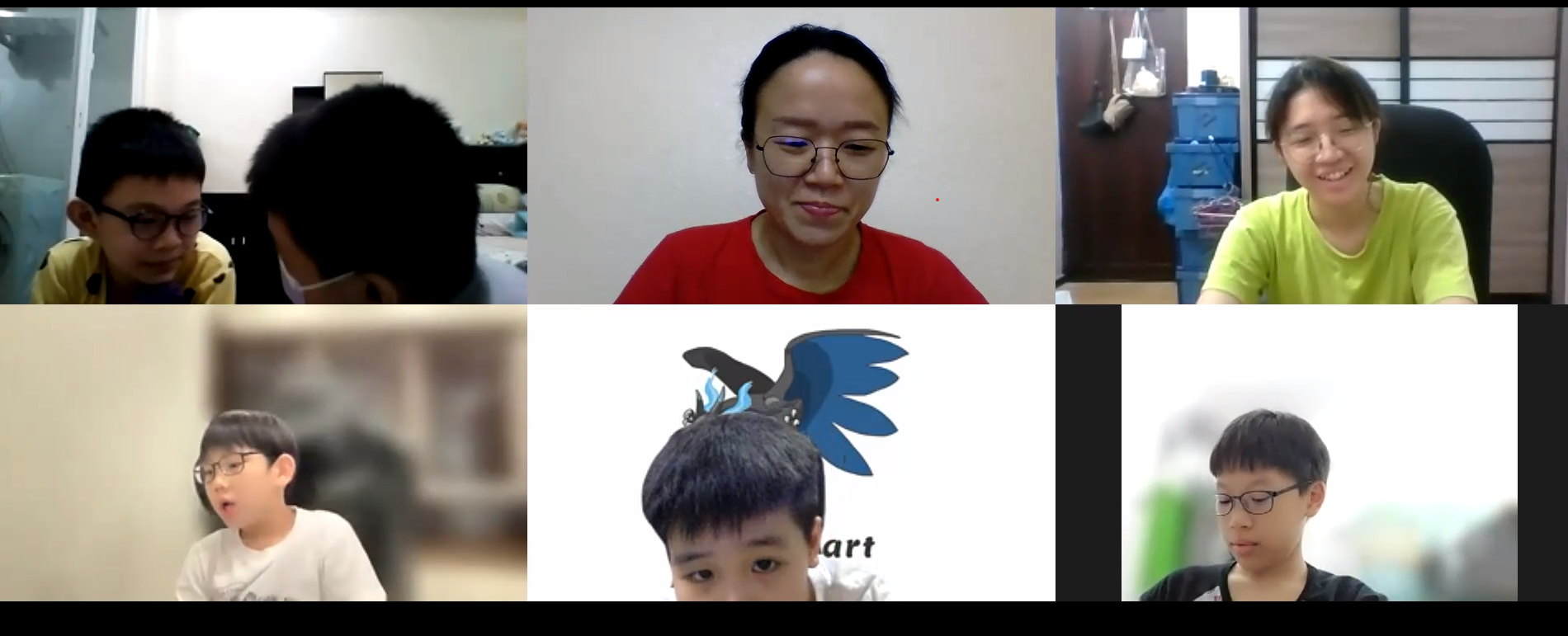}\hfill
    (a)
    \includegraphics[width=.49\textwidth]{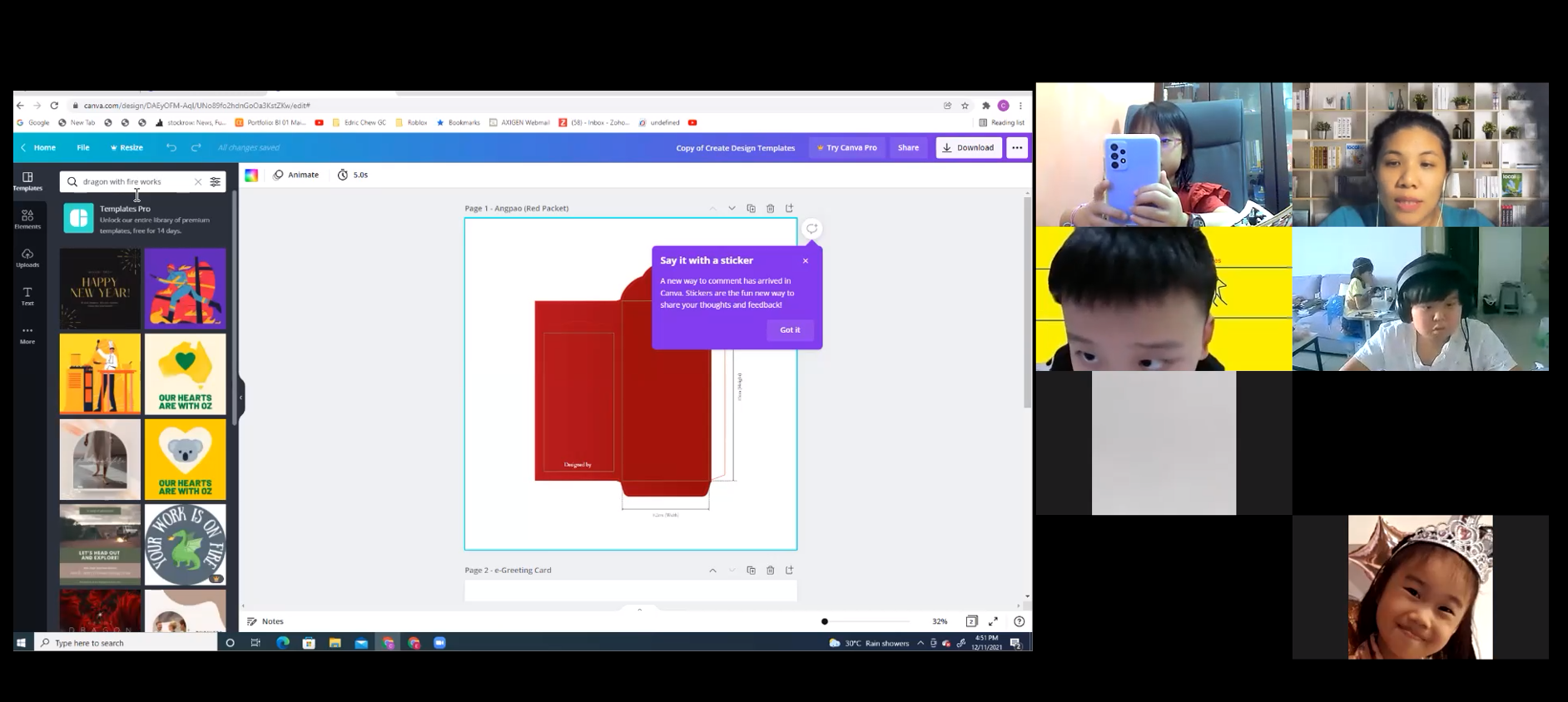}\hfill
    (b)
    \caption{sample frames of raw videos. (a) frame showing only children and facilitators (b) frame showing screen share }
\label{samplesp}
\end{figure}

In order to create a dataset of child micro-expressions, we have obtained recordings of virtual development activities being performed by children at a psychotherapy center. The dataset consists of a large number of unedited video recordings consisting of multiple children as well as psychotherapist(s) engaging in group learning activities. The recordings were made from zoom screen recordings as shown in Figure \ref{samplesp}(a) and occasionally consists of screen shares as well, shown in Figure \ref{samplesp}(b). The recordings are made in 25 frames per second.

In contrast to existing datasets conducted in laboratory settings using pre-determined videos to elicit emotions, the CMED dataset captures naturalistic interactions between children and psychotherapists during group learning activities, providing context-rich data rather than controlled or artificial scenarios. This increases the real-world validity of the dataset. In addition group learning activities encourage various emotional expressions, such as happiness, surprise, disgust, and anger, enabling the dataset to cover a broad spectrum of micro-expressions due to group dynamics, such as peer influence or reactions to authority figures.

\subsection{Data Pre-processing}
In order to reduce the workload of the human labellers and increase the efficiency of the labelling process, certain pre-processing steps are applied to the sample videos prior to them being sent for labelling. The aim of this step is as follows:

\begin{enumerate}
\item Remove sections of videos with screen share.
\item Split the video into segments where each segment consists of only a single subject. Each spatially split segment consists of a 300 $\times$ 300 pixels.
\item Remove parts of the video where the faces of the subjects greatly rotated, at a large distance from the camera or are otherwise obstructed, and thus standardise the samples to a certain extent.
\item Remove any extremely short samples arising following the above
\end{enumerate}

In order to achieve the above, we have created an algorithm that splits videos into chunks in both the spatial and temporal dimensions, where each chunk consists of a single subject continually appearing without obstruction. Algorithm \ref{alg} shows how the frames in the video are iterated over, and when it detects faces, for each detected face, it creates a sample until the face moves out of the bounding area assigned to it.

\begin{algorithm}
\caption{Automated Pre-Processing}
\label{alg}
\begin{algorithmic}[1]
\State Initialize video\_writers[], frame\_counter, no\_face\_count, user\_frame\_list[], user\_frame\_listFlags[], and vidFrames[]\;
\For{frame\_counter in range(total\_frames)}
    \If{frame\_counter \% 60 == 0}
        \State Read the next three frames from the video\;
        \State Convert each frame to grayscale\;
        \State Detect faces in each frame using the face\_cascade classifier\;
        \State Identify the list with the most faces detected among the three frames\;
        \For{face in largest\_list}
            \If{face does not belong to any existing user frame}
                \State Create a new user frame and initialize a video writer for this user\;
            \EndIf
            \If{face belongs to an existing user frame}
                \State Continue writing frames related to that user's video feed\;
                \If{user's face disappears for more than 60 frames}
                    \State Close the video writer and save the video to the output folder\;
                \EndIf
                \If{user's video feed is shorter than 3 seconds}
                    \State Close the video writer and delete the video\;
                \EndIf
            \EndIf
        \EndFor
    \EndIf
\EndFor
\State Release the video capture and all video writers\;
\end{algorithmic}
\end{algorithm}

\begin{figure}[h]
\centering

    \includegraphics[width=.24\textwidth]{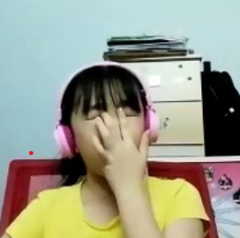}\hfill
    \includegraphics[width=.24\textwidth]{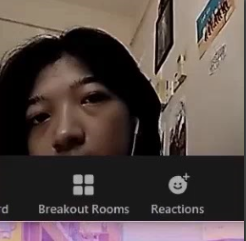}\hfill(a)  \hspace{.23\textwidth}  (b)
    \includegraphics[width=.24\textwidth]{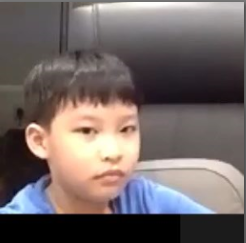}\hfill
    \includegraphics[width=.24\textwidth]{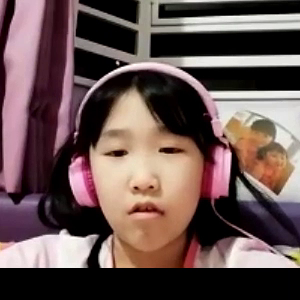}\hfill
    (c)  \hspace{.23\textwidth}  (d)
    \caption{sample frames after automated preprocessing. (a) exclusion criteria - face is covered (b) exclusion criteria -  black bars or parts of other segments present in frame (c) \& (d) minor intrusions }
\label{samplesp2}
\end{figure}

Following the automated sampling process, videos are manually checked to ensure they do not meet any of the following exclusion criteria:

\begin{enumerate}
\item there is a facilitator in the video (i.e. an adult instead of the child); 

\item  there are other people in the background other than the child;

\item  there is more than one child present (since, in certain instances, siblings may be seated together and thus appear in a single sample);

\item  the children are turning, or moving rapidly;

\item  the children are fidgeting, eating, or otherwise covering their face in the majority of the sample.

\item videos may have black bars or parts of other screen segments (other user's screens, etc) in the sample. If the intrusion is significant as in Figure \ref{samplesp2}(b), the sample is excluded. However, if the intrusion is minor, predominantly static and does not obstruct the face, as in Figure \ref{samplesp2}(c) the sample may be retained.

\end{enumerate}

Following the pre-processing, a total of 5684 minutes of video were compiled and sent to clinicans for labelling as discussed in Section III(C).

\subsection{Labeller selection and training to standardize}

In order to obtain the ground truth labels for the dataset, two psychotherapists experienced in child psychotherapy were selected. They were then provided with training using the METT tool which is used to help in the detection of micro-expressions. As part of the training, each labeller was required to undertake the following courses from paulekman.com\cite{EkmanWebsite} which provides extended practice in detecting micro and subtle expressions and is a commonly used training tool in the industry.
\begin{enumerate}

\item Micro Expressions Training Tool
\item Subtle Expressions Training Tool
\item Micro Expressions Profile Training Tool
\item Micro Expressions Intensive Training Tool
\end{enumerate}

This is completed in order to standardize the labelling process of both labellers.  

For the purposes of the dataset, we label micro-expressions as one of 6 emotion classes Happiness, Sadness, Anger, Fear, Disgust, or Surprise. In addition, we scan the videos for segments where the face is not emitting any micro-expressions and classify them as No-ME. The purpose of the No-ME class is to allow for further research into detecting the presence of micro-expressions with verified samples where micro-expressions are not present. Most current models aim to classify given samples to one emotion class; however, in real-world scenarios, micro-expressions may not be present at all times and thus, by incorporating the No-ME class into the dataset, future models would be able to avoid overfitting all samples to an emotion class

\subsection{Labelling Methodology}
The labelling process was split into three stages; the initial labelling stage, a verification stage and an optional re-verification stage. In the initial labelling stage, one labeller is provided with pre-processed videos and is required to note the start and end of the micro-expression within the video (i.e. the onset and offset frame) and are required to select one of the 7 specified classes ( Happiness, Sadness, Anger, Fear, Disgust, Surprise and No-ME). If one of the emotion classes is selected (i.e. any class other than No-ME), the labeller is required to also note the apex frame, where the movement is at its peak.

Once the first stage is complete for a sample, the videos are temporally cropped from the onset to offset frame. These videos are then sent for verification in the second stage. In this stage, the second labeller will review each cropped sample and provide a classification for each of the samples. This classification is provided blindly, without prior knowledge of the label provided in the first stage. The classification provided by the labeller in the first stage is then compared with the classification provided by the labeller in the second stage, and the samples where the classifications are in agreement are selected to be added to the dataset. 

The samples with inter-labeller dis-agreement are then sent to the third, re-verification stage. In the third stage, the labeller in the second stage is provided the samples and asked to reclassify it as either the original label (from the first stage) or the verification label (provided by the same labeller during the second stage) or a different label. In order to maintain the integrity of the process, the labellers are not told which of the labels was from the labelling stage or the verification stage. If the re-verified label matches the original label from the first stage, it is then added into the dataset.

\begin{figure}[h!]
\centering

    \includegraphics[width=.16\textwidth]{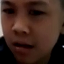}\hfill
    \includegraphics[width=.16\textwidth]{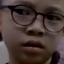}\hfill
    \includegraphics[width=.16\textwidth]{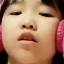}\hfill(a) 
    
    \includegraphics[width=.16\textwidth]{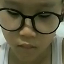}\hfill
    \includegraphics[width=.16\textwidth]{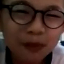}\hfill
    \includegraphics[width=.16\textwidth]{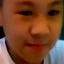}\hfill(b) 
    
    \includegraphics[width=.16\textwidth]{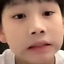}\hfill
    \includegraphics[width=.16\textwidth]{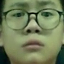}\hfill
    \includegraphics[width=.16\textwidth]{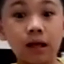}\hfill(c) 
    
    \includegraphics[width=.16\textwidth]{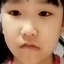}\hfill
    \includegraphics[width=.16\textwidth]{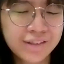}\hfill
    \includegraphics[width=.16\textwidth]{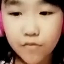}\hfill(d) 
    
    \includegraphics[width=.16\textwidth]{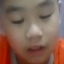}\hfill
    \includegraphics[width=.16\textwidth]{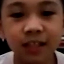}\hfill
    \includegraphics[width=.16\textwidth]{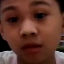}\hfill(e) 
    
    \includegraphics[width=.16\textwidth]{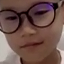}\hfill
    \includegraphics[width=.16\textwidth]{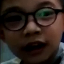}\hfill
    \includegraphics[width=.16\textwidth]{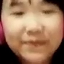}\hfill(f) 
    \caption{Sample apex frames from the dataset (a) anger (b) disgust (c) fear (d) sadness (e) surprise (e) happiness }
\label{samplesfinal}
\end{figure}

\subsection{Dataset Details}

Figure \ref{samplesfinal} illustrates a few apex frames from each emotion class in the dataset. Video samples from 74 different subjects make up the dataset. Table \ref{classdetails} details the number of samples recorded for each micro-expression class, providing an overview of the distribution of samples into the 7 classes. The breakdown of samples per class for each subject is displayed in Table \ref{classdetailsbysubjects}, which adds to our understanding of the dataset's coverage and comprehensiveness by providing insights into how the data were distributed among participants. This subject-wise distribution helps assess dataset diversity and ensure broad representation of emotional variations among different individuals. The number of samples per subject varies significantly, with some subjects contributing over 1,000 samples while others provide only a handful. This natural variability is expected, as individual expressivity and response to stimuli differ across participants. In addition to the 6 class variant of the dataset, we have compiled a dataset for a 3 class variant, similar to the MEGC-2019 guidelines \cite{See2019} where the Anger, Disgust, Fear and Sadness classes are classed as Negative and Happiness is classed as Positive. This is completed in order to reduce the class imbalance in most micro-expression datasets. The number of samples in each class of the re-classified dataset is provided in Table \ref{3classdetails}.

Table \ref{framedetails} presents the maximum, minimum, and average number of frames recorded for the dataset. This draws attention to the sample length variation, which is essential for effectively capturing the transient and transitory character of micro-expressions.

\begin{table}[h!]
    \centering
    \caption{Number of Samples for Each Class}
    \begin{tabular}{|l|c|}
        \hline
        \textbf{Class} & \textbf{Number of Samples} \\
        \hline
        Anger & 114 \\
        \hline
        Disgust & 105 \\
        \hline
        Fear & 19 \\
        \hline
        Sadness & 42 \\
        \hline
        Happiness & 4,855 \\
        \hline
        Surprise & 351 \\
        \hline
        \textbf{Total Emotion Samples} & \textbf{5,486} \\
        \hline
        No Emotion & 5,438 \\
        \hline
        \textbf{Total} & \textbf{10,924} \\
        \hline
    \end{tabular}
    \label{classdetails} 

\end{table}

\begin{table}[h!]
    \centering
    \caption{Number of Samples for Each Class for the Re-Classed Variant of the Dataset}
    \begin{tabular}{|l|c|}
        \hline
        \textbf{Class} & \textbf{Number of Samples} \\
        \hline
        Negative & 280 \\
        \hline
        Positive & 4,855 \\
        \hline
        Surprise & 351 \\
        \hline
        \textbf{Total Emotion Samples} & \textbf{5,486} \\
        \hline
    \end{tabular}
    \label{3classdetails} 

\end{table}
\begin{table}[h!]
    \centering
    \caption{Statistics of Frame Counts}
    \begin{tabular}{|c|c|c|c|}
        \hline
        \textbf{Condition} & \textbf{Max Frames} & \textbf{Min Frames} & \textbf{Avg Frames} \\
        \hline
        6 Emotion Classes & 821 & 4 & 90.63 \\
        \hline
        No Emotion Class& 7680 & 9 & 227.51 \\
        \hline
        Overall & 7680 & 4 & 164.32 \\
        \hline
    \end{tabular}
    \label{framedetails}

\end{table}

\begin{table*}[]

\centering
  \caption{Number of Samples for Each Subject by Class}
    \label{classdetailsbysubjects} 

  \bgroup
\def\arraystretch{0.915}%
\begin{tabular}{|l|l|l|l|l|l|l|l|l|}
\hline
\textbf{Subject} & \textbf{Disgust} & \textbf{Happiness} & \textbf{Anger} & \textbf{Fear} & \textbf{Sadness} & \textbf{Surprise}& \textbf{No-ME}& \textbf{Total}\\
    \hline
Sub\_1  & 12  & 602  & 29  & 2  & 19 & 37  & 691  & 1392  \\ \hline
Sub\_2  & 4   & 298  & 10  & 0  & 8  & 15  & 795  & 1130  \\ \hline
Sub\_3  & 4   & 376  & 9   & 2  & 1  & 41  & 499  & 932   \\ \hline
Sub\_4  & 11  & 513  & 1   & 2  & 4  & 12  & 373  & 916   \\ \hline
Sub\_5  & 2   & 245  & 2   & 2  & 0  & 63  & 377  & 691   \\ \hline
Sub\_6  & 8   & 287  & 5   & 0  & 2  & 37  & 253  & 592   \\ \hline
Sub\_7  & 4   & 135  & 10  & 0  & 0  & 16  & 397  & 562   \\ \hline
Sub\_8  & 11  & 370  & 9   & 1  & 0  & 31  & 129  & 551   \\ \hline
Sub\_9  & 7   & 223  & 2   & 1  & 0  & 12  & 167  & 412   \\ \hline
Sub\_10 & 0   & 309  & 0   & 0  & 0  & 8   & 93   & 410   \\ \hline
Sub\_11 & 9   & 78   & 3   & 0  & 0  & 1   & 265  & 356   \\ \hline
Sub\_12 & 6   & 131  & 5   & 4  & 1  & 16  & 175  & 338   \\ \hline
Sub\_13 & 2   & 238  & 2   & 0  & 2  & 18  & 68   & 330   \\ \hline
Sub\_14 & 3   & 140  & 2   & 0  & 0  & 4   & 147  & 296   \\ \hline
Sub\_15 & 2   & 120  & 0   & 0  & 1  & 5   & 144  & 272   \\ \hline
Sub\_16 & 1   & 157  & 0   & 0  & 0  & 1   & 53   & 212   \\ \hline
Sub\_17 & 5   & 42   & 3   & 0  & 2  & 3   & 130  & 185   \\ \hline
Sub\_18 & 0   & 117  & 0   & 0  & 0  & 0   & 42   & 159   \\ \hline
Sub\_19 & 0   & 45   & 1   & 0  & 0  & 3   & 88   & 137   \\ \hline
Sub\_20 & 0   & 71   & 0   & 0  & 0  & 1   & 35   & 107   \\ \hline
Sub\_21 & 2   & 29   & 0   & 2  & 0  & 0   & 48   & 81    \\ \hline
Sub\_22 & 1   & 20   & 1   & 1  & 1  & 2   & 48   & 74    \\ \hline
Sub\_23 & 7   & 31   & 5   & 0  & 0  & 2   & 28   & 73    \\ \hline
Sub\_24 & 0   & 61   & 0   & 0  & 0  & 11  & 0    & 72    \\ \hline
Sub\_25 & 0   & 6    & 0   & 0  & 0  & 0   & 52   & 58    \\ \hline
Sub\_26 & 0   & 23   & 1   & 0  & 0  & 0   & 15   & 39    \\ \hline
Sub\_27 & 0   & 21   & 1   & 1  & 0  & 1   & 12   & 36    \\ \hline
Sub\_28 & 1   & 5    & 0   & 0  & 0  & 0   & 28   & 34    \\ \hline
Sub\_29 & 1   & 6    & 3   & 1  & 0  & 1   & 20   & 32    \\ \hline
Sub\_30 & 0   & 4    & 4   & 0  & 0  & 0   & 22   & 30    \\ \hline
Sub\_31 & 0   & 23   & 0   & 0  & 0  & 1   & 3    & 27    \\ \hline
Sub\_32 & 1   & 3    & 1   & 0  & 0  & 2   & 18   & 25    \\ \hline
Sub\_33 & 0   & 3    & 0   & 0  & 0  & 0   & 22   & 25    \\ \hline
Sub\_34 & 0   & 8    & 0   & 0  & 0  & 0   & 11   & 19    \\ \hline
Sub\_35 & 0   & 16   & 0   & 0  & 0  & 2   & 0    & 18    \\ \hline
Sub\_36 & 0   & 4    & 0   & 0  & 0  & 0   & 12   & 16    \\ \hline
Sub\_37 & 0   & 4    & 0   & 0  & 0  & 0   & 12   & 16    \\ \hline
Sub\_38 & 1   & 4    & 0   & 0  & 0  & 0   & 11   & 16    \\ \hline
Sub\_39 & 0   & 9    & 0   & 0  & 0  & 1   & 5    & 15    \\ \hline
Sub\_40 & 0   & 11   & 0   & 0  & 0  & 0   & 4    & 15    \\ \hline
Sub\_41 & 0   & 2    & 0   & 0  & 0  & 0   & 12   & 14    \\ \hline
Sub\_42 & 0   & 0    & 0   & 0  & 0  & 0   & 14   & 14    \\ \hline
Sub\_43 & 0   & 1    & 0   & 0  & 0  & 0   & 13   & 14    \\ \hline
Sub\_44 & 0   & 6    & 0   & 0  & 0  & 0   & 7    & 13    \\ \hline
Sub\_45 & 0   & 3    & 0   & 0  & 0  & 1   & 9    & 13    \\ \hline
Sub\_46 & 0   & 1    & 0   & 0  & 0  & 0   & 12   & 13    \\ \hline
Sub\_47 & 0   & 0    & 0   & 0  & 0  & 0   & 12   & 12    \\ \hline
Sub\_48 & 0   & 11   & 0   & 0  & 0  & 0   & 0    & 11    \\ \hline
Sub\_49 & 0   & 3    & 0   & 0  & 0  & 0   & 8    & 11    \\ \hline
Sub\_50 & 0   & 3    & 0   & 0  & 0  & 0   & 6    & 9     \\ \hline
Sub\_51 & 0   & 3    & 0   & 0  & 0  & 0   & 6    & 9     \\ \hline
Sub\_52 & 0   & 0    & 3   & 0  & 0  & 0   & 6    & 9     \\ \hline
Sub\_53 & 0   & 0    & 0   & 0  & 0  & 0   & 9    & 9     \\ \hline
Sub\_54 & 0   & 5    & 0   & 0  & 0  & 0   & 3    & 8     \\ \hline
Sub\_55 & 0   & 1    & 0   & 0  & 0  & 1   & 6    & 8     \\ \hline
Sub\_56 & 0   & 3    & 0   & 0  & 0  & 0   & 4    & 7     \\ \hline
Sub\_57 & 0   & 0    & 1   & 0  & 0  & 0   & 6    & 7     \\ \hline
Sub\_58 & 0   & 6    & 0   & 0  & 0  & 0   & 0    & 6     \\ \hline
Sub\_59 & 0   & 3    & 0   & 0  & 0  & 1   & 1    & 5     \\ \hline
Sub\_60 & 0   & 2    & 0   & 0  & 1  & 0   & 2    & 5     \\ \hline
Sub\_61 & 0   & 4    & 0   & 0  & 0  & 0   & 0    & 4     \\ \hline
Sub\_62 & 0   & 2    & 0   & 0  & 0  & 0   & 1    & 3     \\ \hline
Sub\_63 & 0   & 0    & 0   & 0  & 0  & 1   & 2    & 3     \\ \hline
Sub\_64 & 0   & 0    & 0   & 0  & 0  & 0   & 3    & 3     \\ \hline
Sub\_65 & 0   & 2    & 0   & 0  & 0  & 0   & 0    & 2     \\ \hline
Sub\_66 & 0   & 0    & 0   & 0  & 0  & 0   & 2    & 2     \\ \hline
Sub\_67 & 0   & 0    & 0   & 0  & 0  & 0   & 2    & 2     \\ \hline
Sub\_68 & 0   & 1    & 0   & 0  & 0  & 0   & 0    & 1     \\ \hline
Sub\_69 & 0   & 1    & 0   & 0  & 0  & 0   & 0    & 1     \\ \hline
Sub\_70 & 0   & 1    & 0   & 0  & 0  & 0   & 0    & 1     \\ \hline
Sub\_71 & 0   & 0    & 1   & 0  & 0  & 0   & 0    & 1     \\ \hline
Sub\_72 & 0   & 1    & 0   & 0  & 0  & 0   & 0    & 1     \\ \hline
Sub\_73 & 0   & 1    & 0   & 0  & 0  & 0   & 0    & 1     \\ \hline
Sub\_74 & 0   & 1    & 0   & 0  & 0  & 0   & 0    & 1     \\ \hline
Total   & 105 & 4855 & 114 & 19 & 42 & 351 & 5438 & 10924 \\ \hline
\end{tabular}
\egroup

\end{table*}

\subsection{Feature Exploration}
\label{featureExploration}
In order to identify face movements in the samples and demonstrate the value of the dataset, we compute the average movement of each facial landmark from the onset to apex frames. We extract facial landmarks using a two step process. First, to detect a face the Histogram of Oriented Gradients (HOG) feature is used with a pyramid representation for multi-scale identification and a sliding window for classification using a linear classifier. The second step involves identifying the landmarks using Ensemble of Regression Trees (ERT) as proposed by Kazemi et al. \cite{Kazemi_2014_CVPR} and trained on the 300-W face landmark dataset \cite{SAGONAS20163}. 

Once the landmarks were identified, to make the positions invariant to movements of the entire face, we use five base landmarks illustrated in Fig \ref{fig:roi} (blue) to identify a centroid for the face (using Equations \ref{centeroid} - \ref{centeroid3}). This is then used as the origin from which the distances to the landmarks of the ROI are calculated. All distances are normalized to the width of the face in order to generalize for various face sizes:
\begin{figure}[!h]
    \centering
    \includegraphics[width=0.95\linewidth]{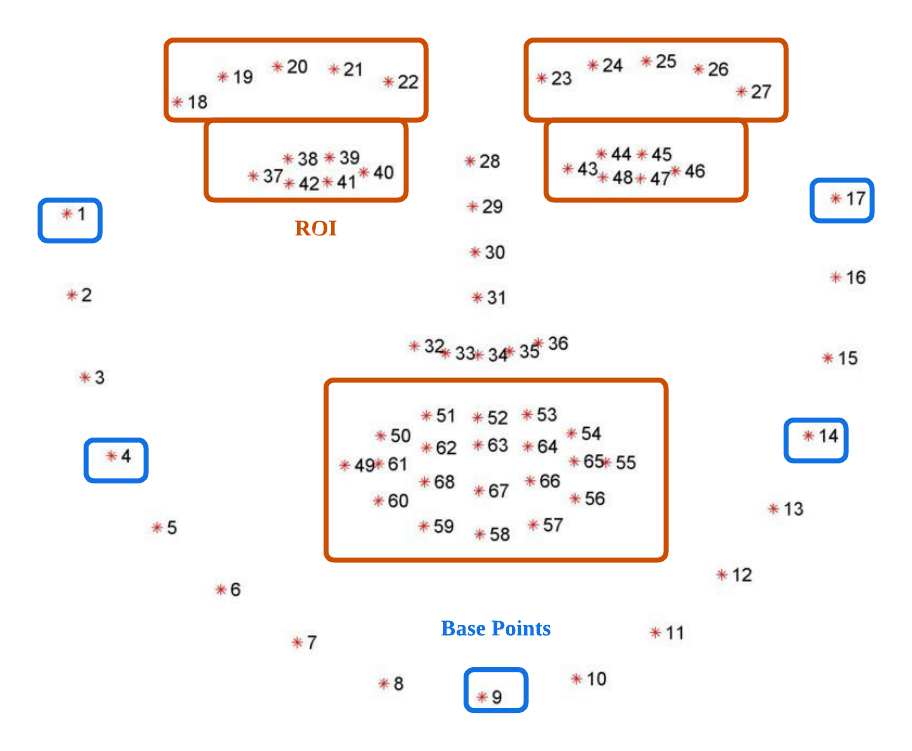}
    \caption{The Region of Interest (ROI) and the base landmarks used for normalization.}
    \label{fig:roi}
\end{figure}

\begin{equation}
\label{centeroid}
  C_x = \frac{1}{c}{\sum^{c}_{i=1}x_c}     
\end{equation}
\begin{equation}
\label{centeroid2}
  C_y = \frac{1}{c}{\sum^{c}_{i=1}y_c}     
\end{equation}
\begin{equation}
\label{centeroid3}
  centroid = (C_x,C_y)    
\end{equation}
where $c$ is the number of base landmarks used in the $x$ and $y$ directions.
\newline

\begin{landscape}
\begin{figure}
\centering

    \includegraphics[width=.30\textwidth]{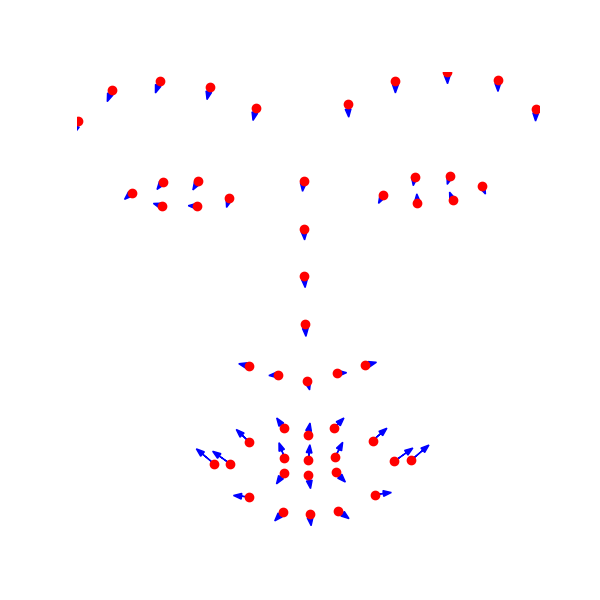}
    \includegraphics[width=.30\textwidth]{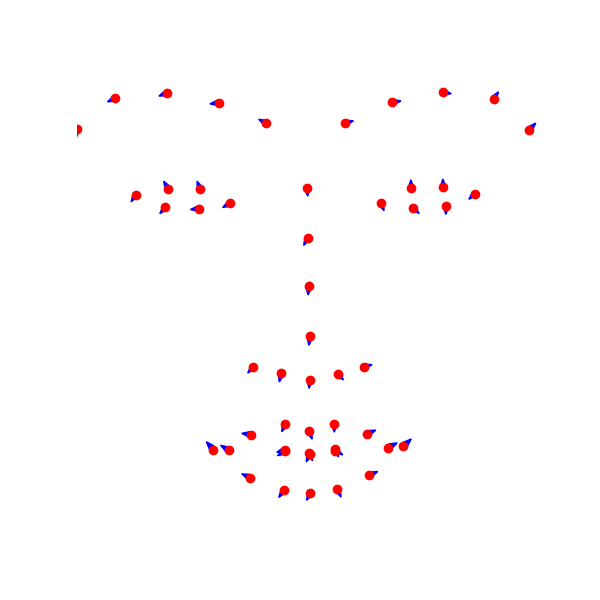}\hfill
    \includegraphics[width=.30\textwidth]{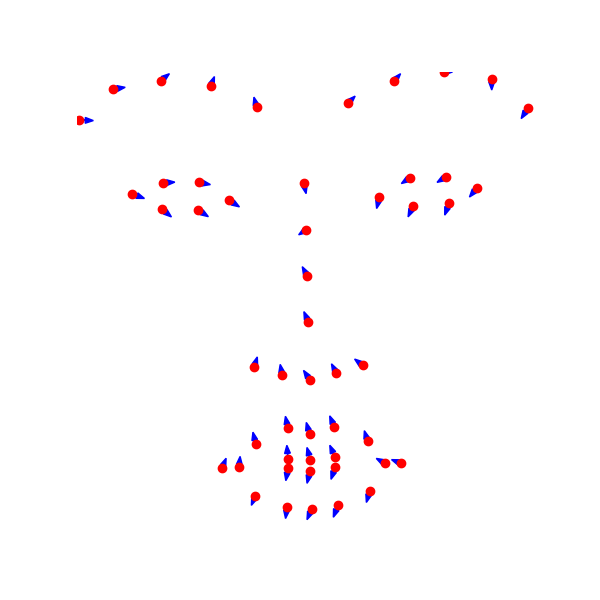}
    \includegraphics[width=.30\textwidth]{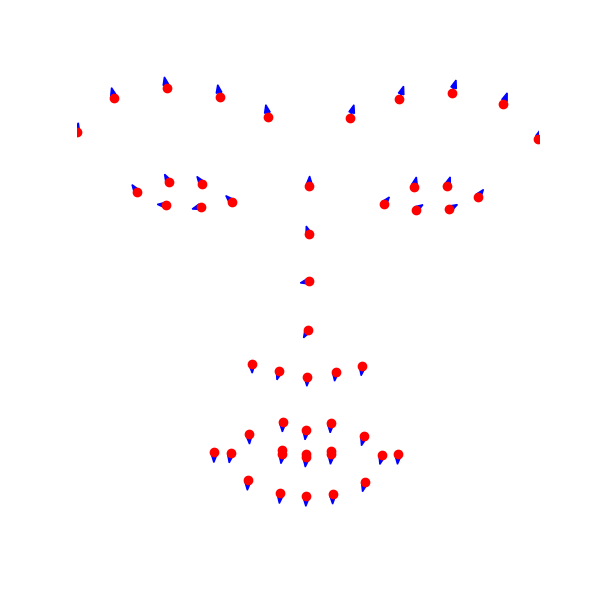}\vspace{-1em} (a)  \hspace{.27\textwidth}  (b)  \hspace{.40\textwidth}  (c)  \hspace{.27\textwidth}  (d) \\

    \includegraphics[width=.30\textwidth]{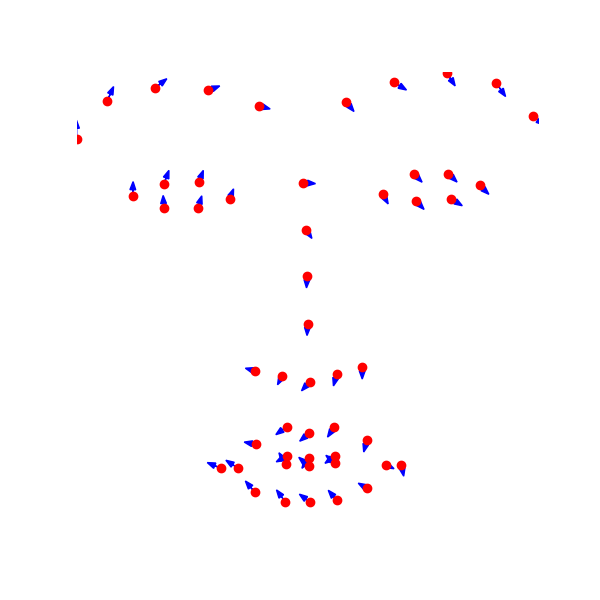}l
    \includegraphics[width=.30\textwidth]{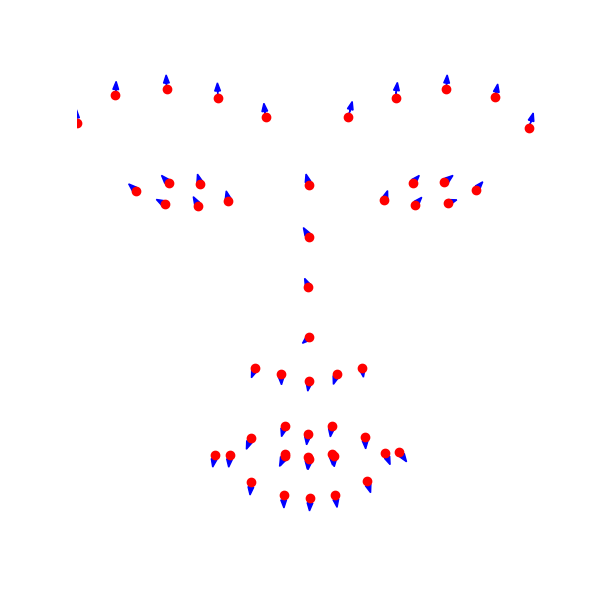}\hfill
    \includegraphics[width=.30\textwidth]{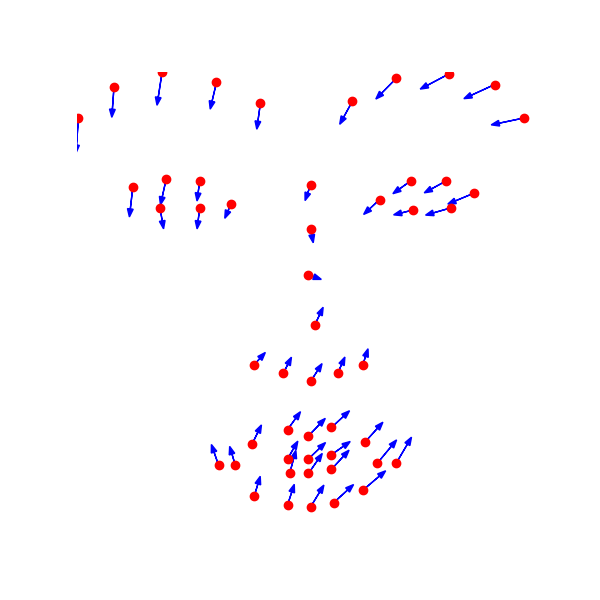}
    \includegraphics[width=.30\textwidth]{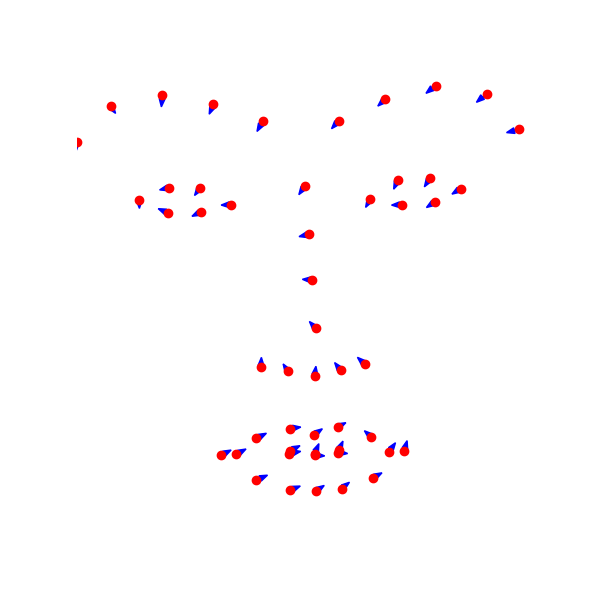}\\\vspace{-1em}(e)  \hspace{.27\textwidth}  (f)  \hspace{.40\textwidth}  (g)  \hspace{.27\textwidth}  (h) \\

    \includegraphics[width=.30\textwidth]{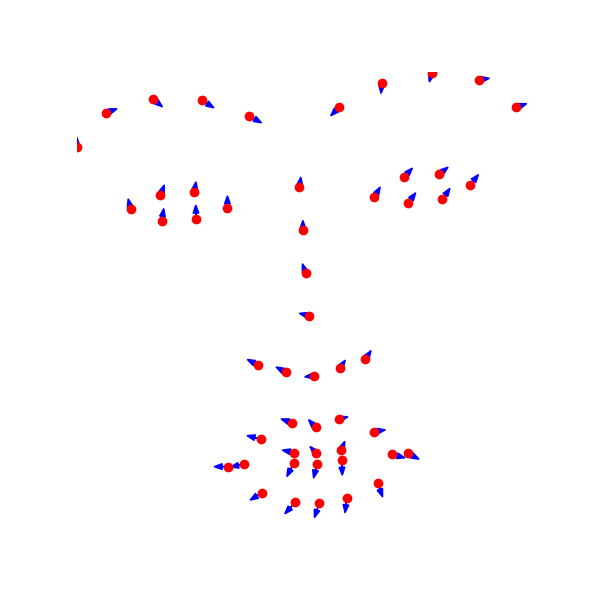}
    \includegraphics[width=.30\textwidth]{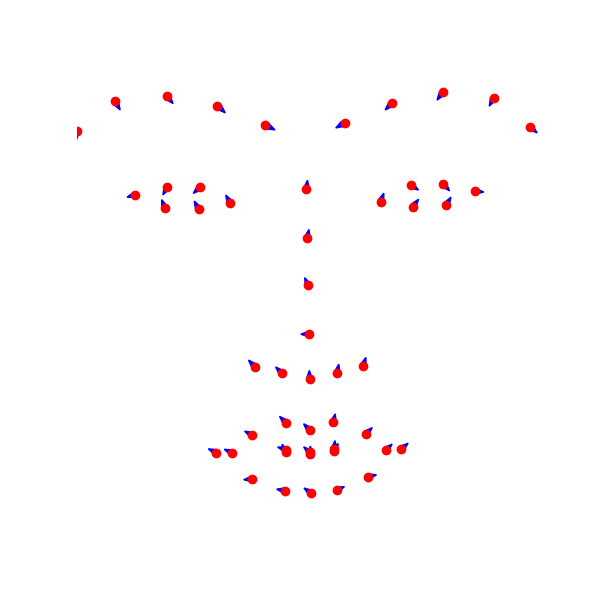}\hfill
    \includegraphics[width=.30\textwidth]{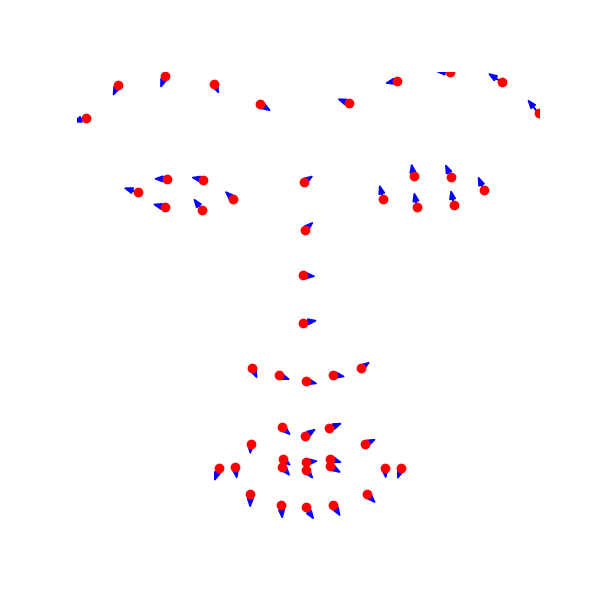}\hfill\\\vspace{-1em}(i)  \hspace{.28\textwidth}  (j)  \hspace{.70\textwidth}  (k)  \\
    
    \caption{Illustration of the average deflection of facial landmarks from the onset to apex frames (a) happiness CMED (b) happiness CASMEII (c) surprise CMED (d) surprise CASMEII (e) sadness CMED (f) sadness CASMEII (g) fear CMED (h) fear CASMEII (i) disgust CMED (j) disgust CASMEII (k) anger CMED }
\label{samplesdeflections}
\end{figure}
\end{landscape}

As illustrated in  Figure \ref{samplesdeflections}, we compare the deflections of landmarks between the child micro-expression dataset and the CASMEII dataset for adult micro-expressions. The illustrations show the landmark point in the onset frame (red dot) and the deflection in the apex frame (blue arrow).

We see that the deflections move in a similar pattern for both datasets. However, there is one key distinction between the child and adult datasets. It is that in the Child ME dataset, the deflections exhibit a larger spread in the $y$-axis for the landmarks compared to the CASMEII dataset, indicating that children may have more exaggerated and less consistent movements for specific emotions. This is most prominently visible in the fear emotion, which shows a much larger deflection in landmarks compared to the other emotions in children.

\section{Evaluation}
In order to establish a baseline for the recognition of micro-expressions in children, we use some of the prominent approaches used in micro-expression recognition of adults. In order to explore a diverse set of approaches, we use Local Binary Patterns from Three Orthogonal Planes
(LBP-TOP) for handcrafted approaches, a VGG16-based Convolution Neural Network (CNN) and a Deep Graph Graph Convolution Neural Network (DGCNN) \cite{ dgcnn} for learning-based approaches.
For each of these approaches, the video samples are first cropped to 224 $\times$ 224 pixels prior to attempting classification.

The conventional LBP method has been proven effective in describing 2D textures. Zhao et al. \cite{zhao2011rotation} introduced the LBP-TOP as an extension of the basic LBP approach, targeting dynamic texture analysis in the spatiotemporal domain. This is achieved by considering three orthogonal planes—XY, XT, and YT—where X and Y are spatial dimensions and T is the temporal dimension as shown in Equation \ref{lbptopeqn}. Each plane captures spatial and temporal information, enabling LBP-TOP to encode the micro-texture changes over time. The onset, apex and offset frames are used for the temporal dimension in the the LBP process.

\begin{equation}
\label{lbptopeqn}
    LBP\_TOP(x,y,t)  = \left\{
    \begin{aligned}
        & \sum_{p=0}^{P_{xy}-1} s(I_{xy}(p) - I_{xy}(c)) \times 2^p, \\
        & \text{(XY-plane)} \\\\
        & \sum_{p=0}^{P_{xt}-1} s(I_{xt}(p) - I_{xt}(c)) \times 2^p,   \\
        &  \text{(XT-plane)} \\\\
        & \sum_{p=0}^{P_{yt}-1} s(I_{yt}(p) - I_{yt}(c)) \times 2^p,  \\
        &  \text{(YT-plane)} 
    \end{aligned} 
    \right.
\end{equation}

where:
\begin{equation}
s(x) =
\begin{cases}
    1, & \text{if } x \geq 0 \\
    0, & \text{if } x < 0
\end{cases}
\end{equation}

\begin{figure}[!h]
    \centering
    \includegraphics[width=0.95\linewidth]{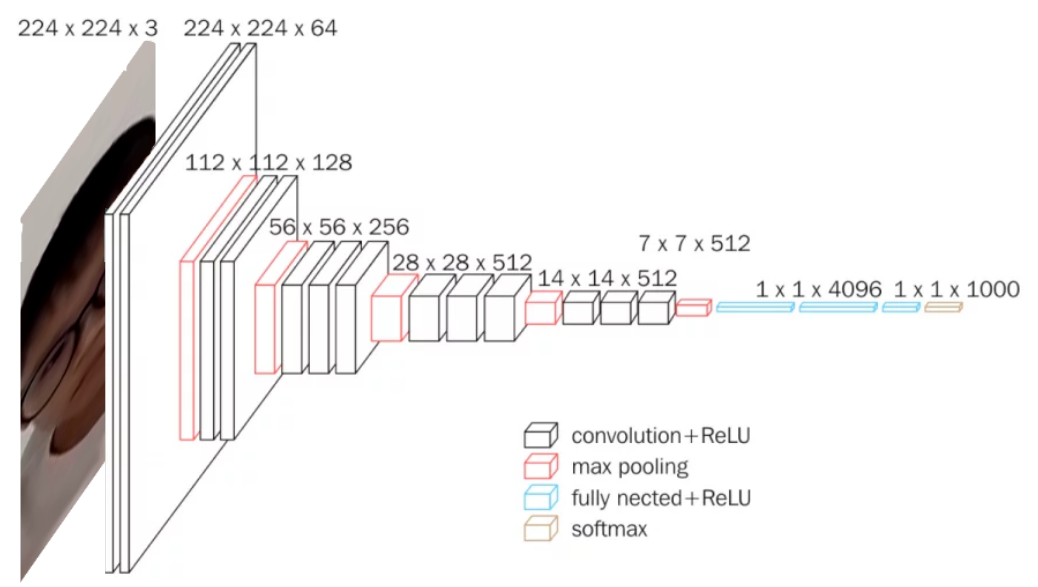}
    \caption{Illustration of the VGG-16 based model used.}
    \label{VGG}
\end{figure}

\begin{itemize}

    \item $\text{I}_{xy}(p)$ , $\text{I}_{xt}(p)$ , $\text{I}_{yt}(p)$ are the intensity values of the neighboring pixels at position $p$.
    \item $\text{I}_{xy}(c)$ ,  $\text{I}_{xt}(c)$ ,  $\text{I}_{yt}(c)$  are the intensity values of the center pixel at  ($x$, $y$, $t$).
    \item $\text{P}_{xy}$ , $\text{P}_{xt}$ , $\text{P}_{yt}$  are the number of sampling points in each plane.
  
\end{itemize}
The second approach used was a VGG-16 Model pre-trained with ImageNet data and transfer-learn using the child micro-expression data. This is an approach that has been successfully used in adult micro-expression recognition and allows us to experiment with the possibility of using a pre-trained model to transfer learn child micro-expression data. The architecture of the model used is illustrated in Figure \ref{VGG}.
Previous studies have shown the importance of apex frames in detecting micro-expressions in adults and that a single frame is able to make predictions on the micro-expression class. Thus to evaluate the importance of the apex frame in child micro-expression recognition, we utilize only the apex frame in this approach.

\begin{figure}[!h]
    \centering
    \includegraphics[width=0.95\linewidth]{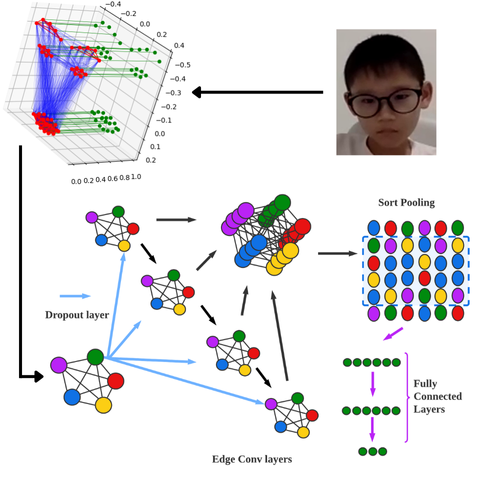}
    \caption{Illustration of how graphs were generated from sample videos and the DGCNN model architecture.}
    \label{DGCNN}
\end{figure}

The third approach uses a graph-based network. We first extract the 68 landmarks as described in section \ref{featureExploration}. Out of the 68 landmarks detected, we use a subset from the Regions of Interest (ROI) depicted with red boxes in Fig. \ref{fig:roi} as it has been demonstrated \cite{nikin}\cite{kumar22b} that the eye, eyebrow, and mouth regions are the most pertinent in distinguishing micro-expressions. Next, we convert the features into a 3D graph with a triplet-of-frames approach where each landmark in the face is connected to the corresponding landmarks in the adjacent key frames. We use a modified version of the DGCNN model initially proposed by Zhang et al. \cite{dgcnn} and adapted for use in facial landmark graphs in \cite{nikin} by using four Edge Convolution (EdgeConv) Layers proposed by Yue Wang et al. \cite{edgeconv} followed by a sortpool and, finally, fully connected layers for a softmax output. This model was previously experimented on for adult micro-expression data and was shown to be robust and time-efficient compared to the previous two approaches. The pipeline of generating graphs and the architecture of the DGCNN model described is illustrated in Figure \ref{DGCNN}.

In order to perform an analysis comparable to state-of-the-art work on micro-expression recognition, we use a leave-one-subject-out as the cross-validation (LOSO-CV) strategy analogous to the work in previous studies \cite{cecv2020} \cite{hong2022}. This method tests the model's performance on all videos of a single subject while training with the data from the remainder of the videos. By repeating the process for each subject, LOSO-CV ensures that all the samples are tested while ensuring that the same subject's samples are not duplicated in the training and test sets.

The overall recognition accuracy (as in equation \ref{acc}), unweighted average recall (UAR) (as in equation \ref{uar}) and F1 score (as in equation \ref{f1}) are used to evaluate the performance
\begin{equation}
\label{f1}
    UF1 = \frac{1}{N} \sum_{i=1}^{N} F1_i
\end{equation}

where:
\begin{itemize}
    \item $N$ is the number of classes.
    \item $F1_i$ is the F1 score for each class $i$, calculated as:
    \begin{equation}
        F1_i = 2 \times \frac{\text{Precision}_i \times \text{Recall}_i}{\text{Precision}_i + \text{Recall}_i}
    \end{equation}
    \item $\text{Precision}_i$ and $\text{Recall}_i$ are defined as:
    \begin{equation}
        \text{Precision}_i = \frac{|\text{TP}_i|}{|\text{TP}_i| + |\text{FP}_i|} \\
        \end{equation}

        \begin{equation}\label{recall}
        \text{Recall}_i = \frac{|\text{TP}_i|}{|\text{TP}_i| + |\text{FN}_i|}
        \end{equation}

\end{itemize}

\begin{equation}
\label{acc}
\text{Accuracy} = \frac{|\text{TP}| + |\text{TN}|}{|\text{TP}| + |\text{TN}| + |\text{FP}| + |\text{FN}|}
\end{equation}

\begin{equation}
\label{uar}
 \text{UAR} = \frac{1}{N} \sum_{i=1}^{N} \text{Recall}_i
\end{equation}

where:
\begin{itemize}
    \item $N$ is the number of classes.
    \item $\text{Recall}_i$ for each class $i$ is calculated as in Equation \ref{recall}
\end{itemize}
\subsection{Micro-expression Spotting}
As a first step we aim to identify if a given sample contains a micro-expression. To that end we classify all micro-expression classes into a single class and perform binary classification with it against the No-ME class. In order to experiment with this, however, it requires us to make a few changes to the VGG-16 and GCN based approaches since the No-ME class does not contain an apex frame. Thus, for the two approaches, a random frame is selected between the onset and offset instead of the apex frame.
\subsection{Micro-expression Recognition}
All of the three approaches were tested using a 6 class configuration for the 6 emotion classes as well as the 3 class variant and the results are shown in Table \ref{resultspreaug}. As shown in Table \ref{resultspreaug}, we can observe a clear performance improvement in deep learning approaches compared to handcrafted approaches, similar to results seen in adult micro-expression studies, due to the limited ability to generalize complex patterns. We can see that the DGCNN model is particularly effective in handling non-Euclidean data, such as facial landmarks or temporal changes in micro-expressions. It models the dynamic interactions between features, leading to robust performance in recognizing subtle micro-expressions. The VGG-16 based CNN approach outperforms traditional methods like LBP-TOP  and the graph based approach, by utilizing its hierarchical feature representations and strong capability in image classification tasks. However, it must be noted that the VGG-16 based method takes significantly longer to make predictions when compared to the DGCNN approach.

\begin{table}[h!]
    \centering
    \caption{Performance Comparison of LBP-TOP, VGG-16, and DGCNN Methods on 6-Class and 3-Class Variants}
    \begin{tabular}{|l|c|c|c|c|}
        \hline
        \textbf{Method} & \textbf{Variant} & \textbf{UF1} & \textbf{UAR} & \textbf{Accuracy} \\
        \hline
        LBP-TOP & 6-Class & 0.1818 & 0.1818 & 0.8641 \\
        \hline
        VGG-16 & 6-Class & 0.2704 & 0.2630 & 0.8893 \\
        \hline
        DGCNN & 6-Class & 0.2184 & 0.2356 & 0.9024 \\
        \hline
        LBP-TOP & 3-Class & 0.3718 & 0.3657 & 0.8439 \\
        \hline
        VGG-16  & 3-Class & 0.5882 & 0.5782 & 0.8913 \\
        \hline
        DGCNN & 3-Class & 0.4463 & 0.4139 & 0.8985 \\
        \hline
    \end{tabular}
    \label{resultspreaug}
\end{table}

\section{Conclusion}
The main aim of this study is to create a collection of labelled spontaneous micro-expression samples from children. To that end, the CMED dataset presents a total of 10,924 samples (including 5,486 micro-expression samples) from 74 different subjects categorized into 6 emotion classes (and no-emotion class). To the best of our knowledge, the CMED dataset is the first of its kind, creating a micro-expression dataset for children. 

The samples were collected using widely commercially available, non-specialized hardware allowing for wider applicability of recognition models arising from the dataset. The use of online screen recordings further improve the real-world applicability of models trained with the CMED dataset as it enables micro-expression recognition in scenarios such as online psychotherapy and online learning environments such as the source of the data.

The results obtained by the learning based methods are promising and they open up
the potential for further improvements to the field.

As a result of the dataset and findings of this study, there is potential for further research into child micro-expressions. While many existing approaches have been built towards identifying micro-expressions in adults, the availability of CMED allows for further work in identifying key differences in child and adult micro-expressions and exploring the potential of creating a unified approach for detecting micro-expressions in adults and children.


%



\section*{Dataset Access}
The CMED dataset is available for research purposes upon request. Interested researchers must complete the CMED Dataset Request Form and sign a Data Access Agreement (DAA) found on the CMED GitHub repository to ensure compliance with privacy and ethical regulations. Requests should be sent via email to Muhammad.FermiPasha@monash.edu along with the completed form. Upon approval, a download link will be provided.

For further details on dataset access and conditions, please refer to the CMED GitHub repository (https://github.com/NikinGeethila/CMED).
\section*{Acknowledgment}

This work was supported by the Malaysian Ministry of Higher Education via the Fundamental Research Grant Scheme (FRGS) project no.: FRGS/1/2020/ICT02/MUSM/03/4

\ifCLASSOPTIONcaptionsoff
  \newpage
\fi



%
\bibliographystyle{IEEEtranN}
\bibliography{bare_jrnl.bbl}


%








\end{document}